\pdfoutput=1

\documentclass[11pt]{article}

\usepackage{acl}

\usepackage{times}
\usepackage{latexsym}
\usepackage{booktabs}
\usepackage{lipsum}
\usepackage{subcaption}
\usepackage{xcolor}
\usepackage{graphicx}
\usepackage{array}
\usepackage{cleveref}
\usepackage{enumitem}
\usepackage[T1]{fontenc}

\usepackage[utf8]{inputenc}
\usepackage{microtype}
\usepackage{xspace}
\newcommand{\dataset}{\emph{MiniPile}\xspace}
\definecolor{orange2}{rgb}{0.95,0.35,0}
 \newcommand{\apr}{\raisebox{0.5ex}{\texttildelow}}

\title{The MiniPile Challenge for Data-Efficient Language Models\\ \textcolor{orange2}{ \normalsize{WARNING: This paper contains NSFW training examples that may be disturbing.}}}

\author{%
Jean Kaddour \\
Centre for Artificial Intelligence\\
University College London \\
\texttt{jean.kaddour.20@ucl.ac.uk} \\
}

\begin{document}
\maketitle
\begin{abstract}The ever-growing diversity of pre-training text corpora has equipped language models with generalization capabilities across various downstream tasks. However, such diverse datasets are often too large for academic budgets; hence, most research on Transformer architectures, training procedures, optimizers, etc. gets conducted on smaller, homogeneous datasets. 

To this end, we present \emph{The MiniPile Challenge}, where one pre-trains a language model on a diverse text corpus containing at most 1M documents. \dataset is a 6GB subset of the deduplicated 825GB \emph{The Pile} \cite{pile} corpus. To curate \dataset, we perform a simple, three-step data filtering process: we (1) infer embeddings for all documents of the Pile, (2) cluster the embedding space using $k$-means, and (3) filter out low-quality clusters. 

To verify \dataset's suitability for language model pre-training, we use it to pre-train a BERT and T5 model, yielding a performance drop of only $1.9\%$/$2.5\%$ on the GLUE and SNI benchmarks compared to the original pre-trained checkpoints trained on 2.6x/745x the amount of data. \dataset is available at \href{https://huggingface.co/datasets/JeanKaddour/minipile}{huggingface.co/datasets/jeankaddour/minipile}.
\end{abstract}

\section{Introduction}
\begin{figure}
    \centering
\includegraphics[width=\columnwidth]{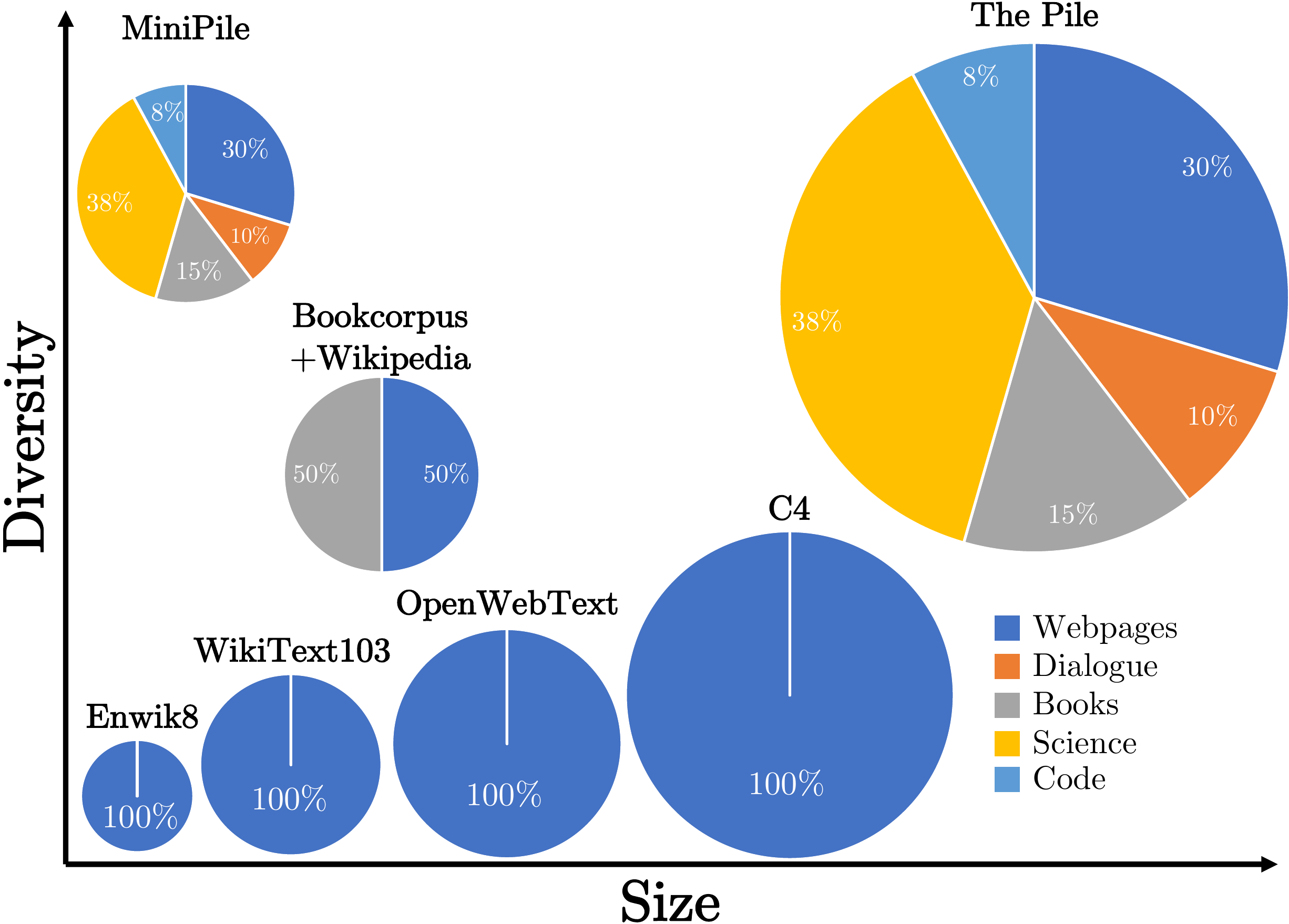}
    \caption{\textbf{MiniPile and Other Pre-Training Datasets.}}
    \label{fig:comparison}
\end{figure}

The Pile \cite{pile} is an 825GB dataset of diverse text, which has gained a lot of popularity in large language model research \cite{lieber2021jurassic,rae2021scaling,borgeaud2022improving,hoffmann2022training,dey2023cerebrasgpt,biderman2023pythia}. It mainly differs from other datasets in its \emph{diversity}: it contains 22 sub-datasets, which can be roughly categorized into webpages, dialogue, books, science, and code \cite{zhao2023survey}, with their proportions shown in \Cref{fig:comparison}. To train language models on diverse datasets of the Pile's size in reasonable training time durations, one requires access to expensive computing resources. 

However, less well-funded ML researchers do not have access to supercomputers and typically fall back on using small-scale, homogeneous datasets unrepresentative of contemporary general-purpose language models. For example, the popular enwik8 \cite{mahoney2011large} / WikiText103 \cite{wikitext} corpora (0.1/1GB large) are still being heavily used for validation of novel research ideas, despite them consisting of only Wikipedia articles and being relatively small. \citet{nagatsuka2022length} show that pre-training a BERT model on Wikitext103 results in GLUE downstream performances much worse than the original BERT model \cite{devlin-etal-2019-bert}, which was trained on a 16GB corpus.  

In this work, we aim to fill in this gap by introducing \dataset, a curated subset of the Pile \cite{pile} that comprises 1 million documents and an uncompressed volume of 6GB. Our goal is to facilitate research on \emph{data-efficient} language model pre-training, joining a broader line of recent work challenging the need for ever-growing computational resources \cite{schwartz2019green,izsak2021train,yao2022nlp,cramming,Nawrot_nanoT5_2023}. 

To curate \dataset and filter out documents we consider harmful or low-quality, we cluster the embedding space of the Pile documents using a state-of-the-art embedding model. Then, we filter out unwanted clusters, with rationales provided in \Cref{sec:excluded_clusters}.
Lastly, we provide first evidence for \dataset being an information-rich pre-training dataset by pre-training a BERT-/T5-Base model on it. After fine-tuning with the GLUE 
 \cite{wang2018glue}/SNI \cite{wang2022supernaturalinstructions} benchmark data, our pre-trained models reach reasonable downstream performances with only small drops compared to models pre-trained on much bigger datasets.

\section{Pruning the Pile}
Our Pile pruning pipeline consists of three steps: (1) document embedding extraction, (2) clustering of embeddings, and (3) human-guided exclusion of unwanted clusters. 

Our starting point is the deduplicated The Pile dump, released by EleutherAI on the HF hub\footnote{\href{https://huggingface.co/datasets/EleutherAI/the_pile_deduplicated}{huggingface.co/datasets/EleutherAI/the\_pile\_deduplicated}}. 

First, we infer embeddings for all documents using \texttt{E5-Large} (\textbf{E}mb\textbf{E}ddings from bidir\textbf{E}ctional \textbf{E}ncoder r\textbf{E}presentations) \cite{wang2022text}, a state-of-the-art text embedding model, which achieves excellent performance on the MTEB benchmark \cite{mtebleaderboard}. 

Second, we cluster the embeddings, motivated by recent work demonstrating clusterability of data subset embeddings \cite{kaddour2020probabilistic,sorscher2022beyond}.
For the clustering, we use batchified $k$-means clustering with the cosine distance between normalized embeddings, $k=220$ ($10$ clusters per the Pile subset), and batch size $16384$. We examined random examples across different clusters and found clear semantic boundaries. For example, we find both clusters matching the high-level categorization from \Cref{fig:comparison} and more fine-grained categories, such as pure mathematics, physics, different programming languages, real estate listings, sports/crime/politics news, etc. 

Third, to decide whether to keep or drop a cluster, we first sort the documents within each cluster by their distance to their assigned centroid \cite{sorscher2022beyond}. Then, a human annotator (the author) judges the data quality based on the  five closest and five most distant examples. We stress that this is a rough estimate of the entire cluster's data, and we may unintentionally exclude some valuable examples; nonetheless, considering the immense size of the Pile and the numerous remaining clusters, we deem this approximation to be sufficiently accurate. 

In preliminary BERT training runs, we also tried selecting only the top-$l$ documents closest to the centroid of their assigned clusters, which one may interpret as excluding hard-to-learn outlier examples \cite{sorscher2022beyond}. However, we observed worse GLUE results than simply randomly subsampling documents within each cluster.

\subsection{Excluded Clusters} \label{sec:excluded_clusters}
\begin{table}[h!]
\centering
\resizebox{\columnwidth}{!}{
\begin{tabular}{p{2cm}p{10cm}p{0.8cm}}
\toprule
\bf Category & \bf Example Text \\
\midrule
Near-Duplicates & "check out our new site makeup addiction add your own caption add your own caption add your own caption add your own caption add your own caption add your own caption add your own caption add your own caption add your own caption add your own caption add your own caption sorry for low quality not sorry for downvote" \\ 
& "check out our new site makeup addiction add your own caption add your own caption add your own caption add your own caption add your own caption add your own caption add your own caption add your own caption add your own caption add your own caption add your own caption want more upvotes? be more funny"  \\ \midrule
Pornography & fuck anal movie adult swinger party melbourne nifty erotics icarly tighter the first inch or so, loosens up beyond that point. actually feels just very slightly warmer. big beautiful ebony keisha grey takes an anal p busty natasha nice gets ass indian teen gangbang publisher [...] \\ \midrule 
Navigation Bars & search open menu close menu pc  mobile windows mac linux android iphone and ipad internet security programming lifestyle technology news entertainment productivity creative gaming social media hardware technology explained buying guides smart home diy product reviews free ebooks giveaways top lists about about makeuseof newsletter advertise privacy jobs chats facebook facebook facebook facebook search for :. jump tosections of this pageaccessibility helppress alt + / to open this menuremoveto [...] \\ \midrule
Product specifications & related products super light, starting at just 3. 0 lbsultra thin - just 14. 5mm at its thinnestpremium processing to help you multitaskinnovative ro... tating sound bar for sound you can feelbrighter display with 4k clarity \& imporoved hinge technology read more the thinkpad a285 is a powerful 12. 5 - inch enterprise laptop that has everything you need to get the job done. the latest amd ryzen... \u2122 pro processing and radeon\u2122 vega graphics make multitasking a cinch. biometric and encryption security protect critical... read more asus x540sa 15. 6 \" [...]   \\ \midrule
Long lists of named entities& tag : blogger. com, 1999 : blog - 6954607999061779677thu, 26 apr 2018 09 : 41 : 52 + 0000mp3videoindieeminemnewstop 10unknown artistsdarius ruckerlinkin parkradioac / dcb. o. b hayley williamsbeyonceblack eyed peasbruce driscollbruno marschitlinscolette carrdutch tha kiddakota fanning kristen stewarteaston corbinedward mayaflo rida david guettafugazigeorge michaelgeorgie jamesguns n'roseshot chelle raeivan howardjosh turnerjustin bieberkenny chesneykeshakid cudile louplil [...] \\
\bottomrule
\end{tabular}}
\caption{\textbf{Examples of Excluded Clusters.}}
\label{tab:bad_clusters}
\end{table}
We exclude 38 out of the 220 clusters, with examples of them shown in \Cref{tab:bad_clusters}. The rationales for excluding such clusters are as follows:
\begin{itemize}[leftmargin=*]
    \item \emph{Near-duplicate} documents will contain repetitions, which have been shown to degrade model performance \cite{lee2021deduplicating,hernandez2022scaling,abbas2023semdedup}.
    \item \emph{Pornography} may contain sexist spurious correlations and enforce racial/social stereotypes \cite{birhane2021multimodal,weidinger2021ethical}.   
    \item \emph{Webpage navigation bars/product specifications/long named entity lists} entail long-tail knowledge, which is challenging to learn even for large language models up to 176B parameters \cite{kandpal2022large}.
\end{itemize}

\subsection{\dataset Statistics}
\begin{itemize}[leftmargin=*]
\setlength\itemsep{0pt}
    \item 1M/500/10k training/validation/test examples
    \item \apr 6/3GB un-/compressed space requirements
    \item Vocab size: $32309614$
    \item Median document length: $294$
    \item Longest document length: $929633$
\end{itemize}

\section{Experiments} \label{sec:results}
The primary goal of our experiments is to verify that \dataset is information-rich enough for pre-training a language model, which reaches reasonable fine-tuning performances on standard downstream task benchmarks. We evaluate our pre-trained models on the General Language Understanding Evaluation (GLUE) \cite{wang2018glue} and Super-Natural-Instructions (SNI) \cite{wang2022supernaturalinstructions} benchmarks. 
\begin{table}[]
\centering
\begin{tabular}{@{}ccc@{}}
\toprule
\textbf{Model} & \textbf{Pre-Training} & \textbf{Fine-Tuning} \\ \midrule
BERT-Base                  &  54h                 &  3h \\
T5v1.1-Base                    & 21h & 2h      
\\ \bottomrule
\end{tabular}
\caption{\textbf{Wall-Clock Times} for our experiments using a single NVIDIA RTX 3090 GPU.}
\label{tab:wct}
\end{table}

We run all experiments on a machine with a single NVIDIA RTX 3090 GPU and highlight the wall-clock times in \Cref{tab:wct}. 

As a reference point for comparability, we list the performance obtained by fine-tuning a publicly available checkpoint of the same model architecture but trained on more data, following the same fine-tuning protocol. We emphasize that our goal is not to attain state-of-the-art performance on GLUE/SNI; specifically for these downstream benchmarks, data selected from target distribution \cite{dsir} could be better suited. For example, \citet{cramming} and \citet{Nawrot_nanoT5_2023} reach downstream performances slightly better than ours, using randomly sampled subsets of C4 \cite{raffel2020exploring}. 

\subsection{BERT-style Encoder-Only Masked Language Modeling}

We pre-train a BERT-Base \cite{devlin-etal-2019-bert} model using a masked language modeling (MLM) objective. We adopt the Cramming training recipe \cite{cramming} without further data filtering and use the WordPiece tokenizer with vocabulary size $2^15$, Adam optimizer \cite{adam}, $\beta_1=0.9, \beta_2=0.98, \epsilon=10^{-12}$, weight decay of $0.01$ \cite{adamw}, one-cycle schedule \cite{smith2018superconvergence} with peak learning rate $0.001$, gradient clipping of $0.5$, progressive batch size from $128$ to $4096$ with a linear increase over the course of training up to $300k$ steps, no warmup, $800k$ total training steps, and weight averaging of the $k=5$ latest checkpoints and $1k$ steps distance between them \cite{kaddour2022stop}. 

\subsection{T5-style Encoder-Decoder Span Corruption}
We pre-train a T5v1.1-Base \cite{raffel2020exploring,shazeer2020glu} model using the original span-corrupting MLM objective and SentencePiece \cite{kudo2018sentencepiece} tokenizer. We mostly follow \cite{Nawrot_nanoT5_2023} and use the AdamW optimizer \cite{adamw} with matrix-wise LR scaling by its root mean square (RMS), base learning rate $0.02$, no weight decay, cosine schedule with final of $1e-5$ \cite{cosine_decay}, gradient clipping of $1.0$, batch size of $288$, $10k$ warmup steps, $65536$ total training steps, and weight averaging of the $k=5$ latest checkpoints and $1k$ steps distance between them \cite{kaddour2022stop}. 

\subsection{Discussion}
\begin{table*}[t]
    \centering
\resizebox{\textwidth}{!}{
\begin{tabular}{l|cccccccc|c}
\toprule
 Model & MNLI/-MM & SST-2 & STSB & RTE & QNLI & QQP & MRPC & CoLA & GLUE (Avg.) \\
\midrule
BERT-base \cite{cramming} & 83.2/83.4 & \textbf{91.9} & \textbf{86.7} & \textbf{59.2} & \textbf{90.6} & \textbf{87.7} & \textbf{89.3} & \textbf{56.5} & \textbf{80.9} \\
BERT (\dataset) & \textbf{83.4}/\textbf{84.0} & 91.1 & 83.3 & 58.5 & 90.3 & 87.4 & 88.2 & 45.0 & 79.0 \\
\bottomrule
\end{tabular}}
    \caption{\textbf{GLUE-dev performances} of BERT-base with results provided by \citet{cramming}, and our model pre-trained only on \dataset. 
    }
    \label{tab:glue}
\end{table*}

\begin{table}[t]
    \centering
\resizebox{0.8\columnwidth}{!}{
\begin{tabular}{l|ccc}
\toprule
 Model &  Dev & Test  \\
\midrule
T5v1.1-Base \cite{Nawrot_nanoT5_2023}  & - & 41.0 \\
T5v1.1-Base (\dataset)  & 26.21 & 38.47  \\
\bottomrule
\end{tabular}}
    \caption{\textbf{SNI performances} of baseline T5v1.1-Base \cite{Nawrot_nanoT5_2023} and our model pre-trained only on \dataset. 
    }
    \label{tab:sni}
\end{table}

\begin{table}[t]
    \centering
\begin{tabular}{l|ccc}
\toprule
 Loss &  Dev & Test  \\
\midrule
BERT-Base MLM & 2.01 & 1.98\\
T5v1.1-Base Span-MLM & 1.72 & 1.67  \\
\bottomrule
\end{tabular}
    \caption{\textbf{Model performances} averaged across the \dataset dev and test set. 
    }
    \label{tab:minipile}
\end{table}

\Cref{tab:glue,tab:sni} show the results compared against the publicly available checkpoints trained on $2.6$x/$745$x the amount of data, where we took the numbers from \cite{cramming} and \cite{Nawrot_nanoT5_2023}, respectively. We observe minor reductions in final downstream performance and conjecture that \dataset is a well-suited pre-training corpus for common downstream benchmarks. For future reference, \Cref{tab:minipile} includes the performances of the pre-trained models on \dataset's dev and test set.

\section{Related Work}

\paragraph{Pre-Training Datasets}
Various subsets of Wikipedia dumps have been used in language modeling papers, e.g., \emph{enwik8} \cite{mahoney2011large}, or \emph{WikiText} \cite{wikitext}. \emph{Bookcorpus} \cite{bookcorpus_2} contains >$7$k unpublished books with long stretches of contiguous text. \emph{OpenWebText} \cite{Gokaslan2019OpenWeb} and \emph{C4} \cite{raffel2020exploring} contain text from crawled webpages. Concurrent with this work, \citet{babylm} recently announced the \emph{BabyLM challenge}, with under 100M words of transcribed speech. 
In contrast to \emph{the Pile} and \dataset, these corpora contain less-diverse text.

\paragraph{Data Quality}
The quality of datasets has been questioned in various works, especially in the context of massive collections of web-crawled data \cite{paullada2021data,kreutzer2022quality}. Potential issues with such include token repetitions \cite{lee2021deduplicating,hernandez2022scaling,abbas2023semdedup}, misogyny, pornography, and
malignant stereotypes \cite{birhane2021multimodal,biderman2021pitfalls}, benchmark data contamination \cite{brown2020language,dodge2021documenting}, 
spurious correlations \cite{survey,lynch2022evaluating,lynch2023spawrious}, diluted robustness due to data mixing
\cite{nguyen2022quality} and potentially sensitive information \cite{henderson2022pile}.

\section{Future Work}
We hope to see \dataset accelerating data-efficient language model research, e.g. structurally different architectures \cite{fu2023hungry,zhu2023spikegpt,lan2023copy}, pre-training schemes \cite{rust2022language,min2022nonparametric}, optimizers schemes \cite{anil2021scalable,flatminima,kaddour2022stop}, differential privacy \cite{anil2021large,basu2021benchmarking}, mechanistic interpretability \cite{elhage2021mathematical,olsson2022incontext,belrose2023eliciting}, etc.

\section*{Acknowledgements} I thank Matt Kusner and Ari Morcos for their advice on using the $k$-means algorithm on text embeddings and Jonas Geiping for guidance on BERT pre-training.

\bibliography{anthology,custom}
\bibliographystyle{plainnat}

\end{document}